\def\year{2021}\relax
\documentclass[letterpaper]{article} 
\usepackage{aaai21}  
\usepackage{times}  
\usepackage{helvet} 
\usepackage{courier}  
\usepackage[hyphens]{url}  
\usepackage{graphicx} 
\usepackage{natbib}  
\usepackage{caption} 
\usepackage{amssymb}
\usepackage{multirow}
\usepackage[colorlinks,
            linkcolor=red,
            anchorcolor=blue,
            citecolor=black]{hyperref}
\title{PC-RGNN: Point Cloud Completion and Graph Neural Network for
\\3D Object Detection
}
\author{
    Yanan Zhang\textsuperscript{\rm 1, 2, 3}
    Di Huang\textsuperscript{\rm 1, 2, 3}\thanks{Corresponding author: Di Huang.}
    Yunhong Wang\textsuperscript{\rm 1, 3}
    \\
}
\affiliations{
    \textsuperscript{\rm 1}Beijing Advanced Innovation Center for Big Data and Brain Computing, Beihang University\\
    \textsuperscript{\rm 2}State Key Laboratory of Software Development Environment, Beihang University\\
    \textsuperscript{\rm 3}School of Computer Science and Engineering, Beihang University, Beijing 100191, China\\
    \{zhangyanan, dhuang, yhwang\}@buaa.edu.cn
}
\begin{document}
\maketitle

\begin{abstract}
LiDAR-based 3D object detection is an important task for autonomous driving and current approaches suffer from sparse and partial point clouds of distant and occluded objects. In this paper, we propose a novel two-stage approach, namely PC-RGNN, dealing with such challenges by two specific solutions. On the one hand, we introduce a point cloud completion module to recover high-quality proposals of dense points and entire views with original structures preserved. On the other hand, a graph neural network module is designed, which comprehensively captures relations among points through a local-global attention mechanism as well as multi-scale graph based context aggregation, substantially strengthening encoded features. Extensive experiments on the KITTI benchmark show that the proposed approach outperforms the previous state-of-the-art baselines by remarkable margins, highlighting its effectiveness.
\end{abstract}
\section{Introduction}
3D object detection in point clouds is eagerly in demand in autonomous driving, and LiDAR laser scanners are the most common instruments to collect such data. Compared to 2D images, LiDAR point clouds convey real 3D geometric structures and spatial locations of objects and are less sensitive to illumination variations, which enables more reliable detection results.

In recent years, several approaches have been proposed for 3D object detection and they follow either the one-stage framework or the two-stage one as in the 2D domain, where how to learn effective shape features is an essential issue. For instance, MV3D \cite{chen2017multi} and AVOD \cite{ku2018joint} transform point clouds to bird's eye view or front view as initial representation and apply 2D convolutions for feature map computation; VoxelNet \cite{zhou2018voxelnet} and SECOND \cite{yan2018second} voxelize the 3D space into regular cells and employ 3D convolutions to extract features; F-Pointnet \cite{qi2018frustum}  and PointRCNN \cite{shi2019pointrcnn} take raw point clouds as input and encode features by  PointNets \cite{qi2017pointnet,qi2017pointnet++}.

\begin{figure}[t]
\centering
\includegraphics[width=0.95\columnwidth]{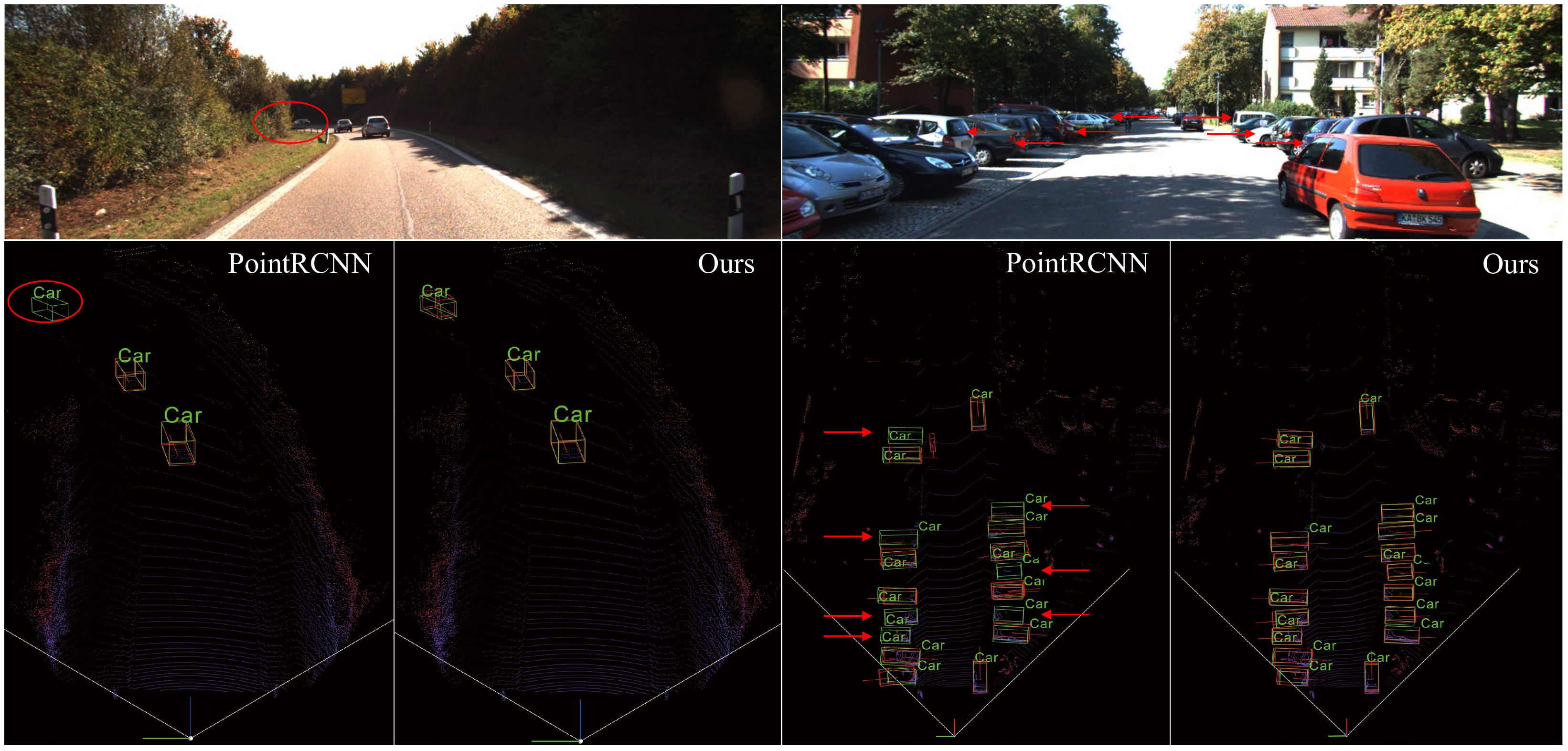} 
\caption{Illustration of the two major challenges in LiDAR-based 3D object detection (best viewed with zoom-in). The left case shows a sparse point cloud for a car far away, while the right case shows the extremely incomplete point clouds for occluded cars. The red and green boxes indicate the predicted results and ground-truths respectively.}
\label{fig1}
\end{figure}

Those approaches indeed show great potentials and have consistently improved the performance of major benchmarks. Unfortunately, they tend to fail in the presence of low-quality input point clouds, \emph{i.e.} sparse and partial data due to distant and occluded objects, which often occur in the real world. As illustrated in Fig. \ref{fig1},
\begin{figure*}[t]
\centering
\includegraphics[width=0.97\textwidth]{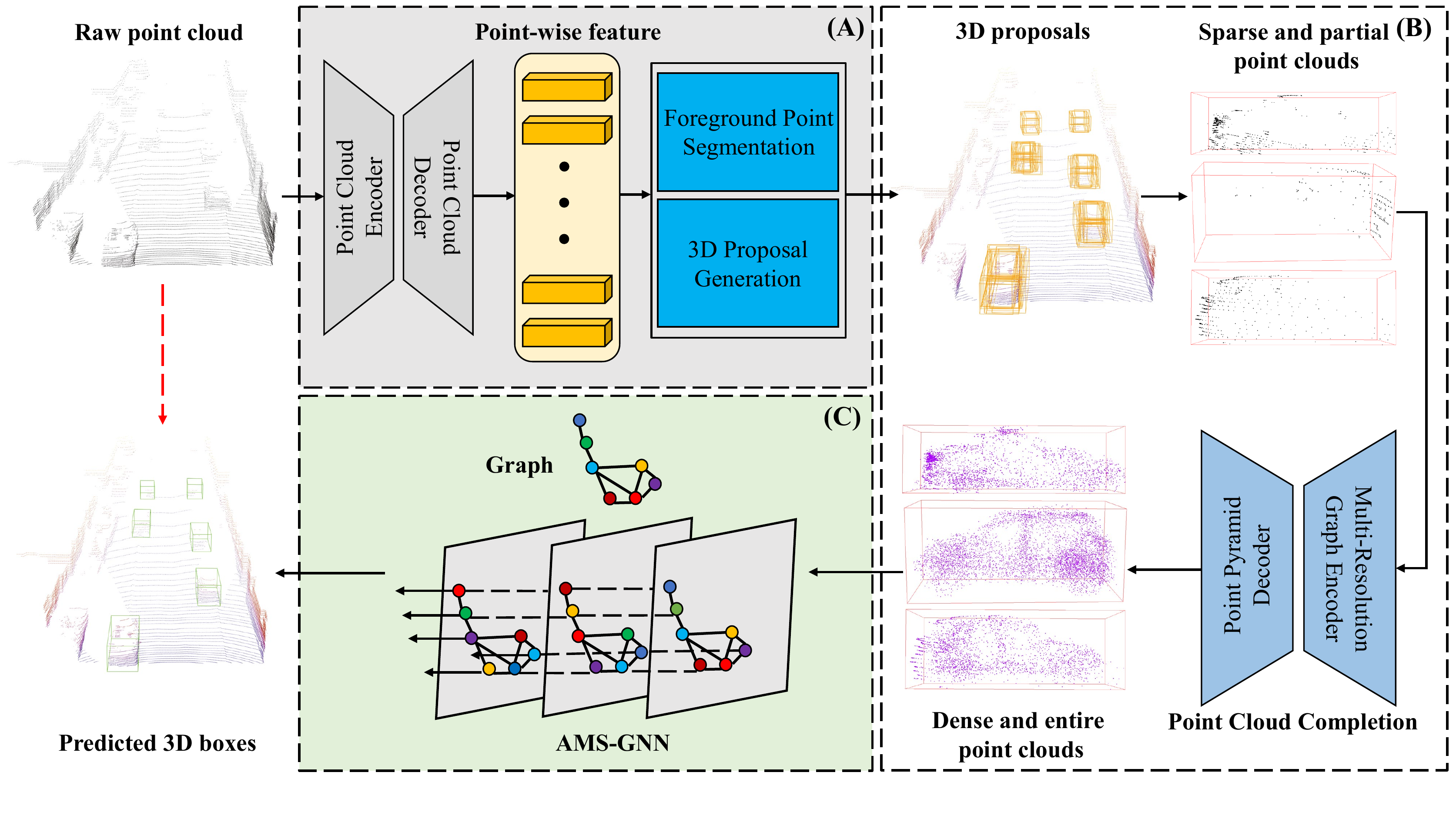} 
\caption{Framework overview. The whole PC-RGNN network consists of three main modules: (A) 3D proposal generation, (B) point cloud completion, and (C) attention based multi-scale GNN representation.}
\label{fig2}
\end{figure*}
PointRCNN, the state of the art representative, misses many objects marked with the red ellipse and arrows for long-distance and severe occlusions. Such a limitation derives from two main aspects: (1) point cloud representation in current 3D object detectors does not work well on large variations in sampling density and sampling integrity and (2) the PointNet-like networks adopted as the backbones in the leading 3D object detectors are not so powerful which make insufficient use of given point clouds.

Motivated by the analysis above, this paper proposes a novel two-stage approach, named PC-RGNN, to 3D object detection from LiDAR based point clouds. Speciﬁcally, the 3D proposal generation (3D PG) module first suggests bounding box candidates in a bottom-up manner via segmenting the whole scene into foreground and background. Since objects are usually sparsely and partially sampled, a point cloud completion (PC) module is then introduced to recover high-quality proposals with dense points and entire views. Further, we model each refined proposal as a graph and design a graph neural network module, called AMS-GNN, to capture its shape characteristics. AMS-GNN aggregates contextual clues by combining multi-scale graphs and learns different weights of neighboring nodes through a local-global attention mechanism, and geometric relations among points can thus be comprehensively exploited, leading to enhanced features for decision making. Thanks to these modules, PC-RGNN reaches very competitive scores on the KITTI database, and in particular, it facilitates the detection of 3D objects in very difficult scenes, as depicted in Fig. \ref{fig1}.

In summary, the main contributions of this paper are:

\begin{itemize}
  \item We highlight the challenge of low-quality input point clouds in LiDAR-based 3D object detection and propose a novel two-stage detection approach (PC-RGNN), which significantly boosts the performance.
  \item We present a new point cloud completion (PC) module to improve proposals of sparse and partial points. To the best of our knowledge, this is the first study that considers point cloud refinement in 3D object detection.
  \item We design a new graph neural network (AMS-GNN) module, which strengths the structure features by encoding geometric relations among points through attention based multi-scale graph aggregation.
\end{itemize}
\section{Related Work}
This section briefly reviews the major approaches to 3D object detection as well as the ones of point cloud completion.

\textbf{Grid-based detectors.} A number of methods initially convert point clouds to regular grids by projecting them to the planes of specific views \cite{chen2017multi,engelcke2017vote3deep,ku2018joint,liang2018deep,liang2019multi} or subdividing them to equally distributed voxels \cite{wang2015voting,zhou2018voxelnet,yan2018second,lang2019pointpillars} so that they can be processed by 2D or 3D CNNs to compute detection features. Although grid-based methods are generally straightforward and efﬁcient, they inevitably incur much information loss and thus limit the performance because of the coarse quantization process.

\textbf{Point-based detectors.} Many methods directly take the raw unordered and irregular points as input and apply point cloud deep learning networks, such as PointNet \cite{qi2017pointnet} and PonintNet++ \cite{qi2017pointnet++}, to encode structure features for detection \cite{qi2018frustum,chen2019fast,qi2019deep,shi2019pointrcnn,yang2019std,shi2020points,yang20203dssd}. These methods outperform grid-based ones. However, without explicit modeling of point relations, they are not so competent at discriminative geometry representation.

\textbf{Graph-based detectors.} Recently, inspired by the success of graph convolutions in point cloud segmentation and classiﬁcation tasks, Point-GNN \cite{shi2020point} investigates the graph neural network to extract shape features for 3D object detection, which proves a promising way. Nevertheless, in their model, each node is regarded to equally contribute in local and global representation and the single-scale graph does not make full use of the contextual information. Both the facts leave space for improvement.

\textbf{Point cloud completion methods.} Point cloud completion aims to estimate entire 3D shapes from partial point cloud inputs. L-GAN \cite{achlioptas2018learning} introduces the first deep learning model with an Encoder-Decoder architecture. PCN \cite{yuan2018pcn} presents a coarse-to-ﬁne procedure to synthesize dense and complete data by a specially designed decoder. RL-GAN-Net \cite{sarmad2019rl} proposes a reinforcement learning agent controlled GAN to speed up the inference phase. PF-Net \cite{huang2020pf} hierarchically recovers the point cloud by a feature-point based multi-scale generation network. Different from the research aforementioned, for the first time, point cloud completion is attempted to ameliorate LiDAR-based 3D object detection, as objects are often very far away or seriously occluded, leading to sparse and partial sampling.
\begin{figure*}[t]
\centering
\includegraphics[width=0.97\textwidth]{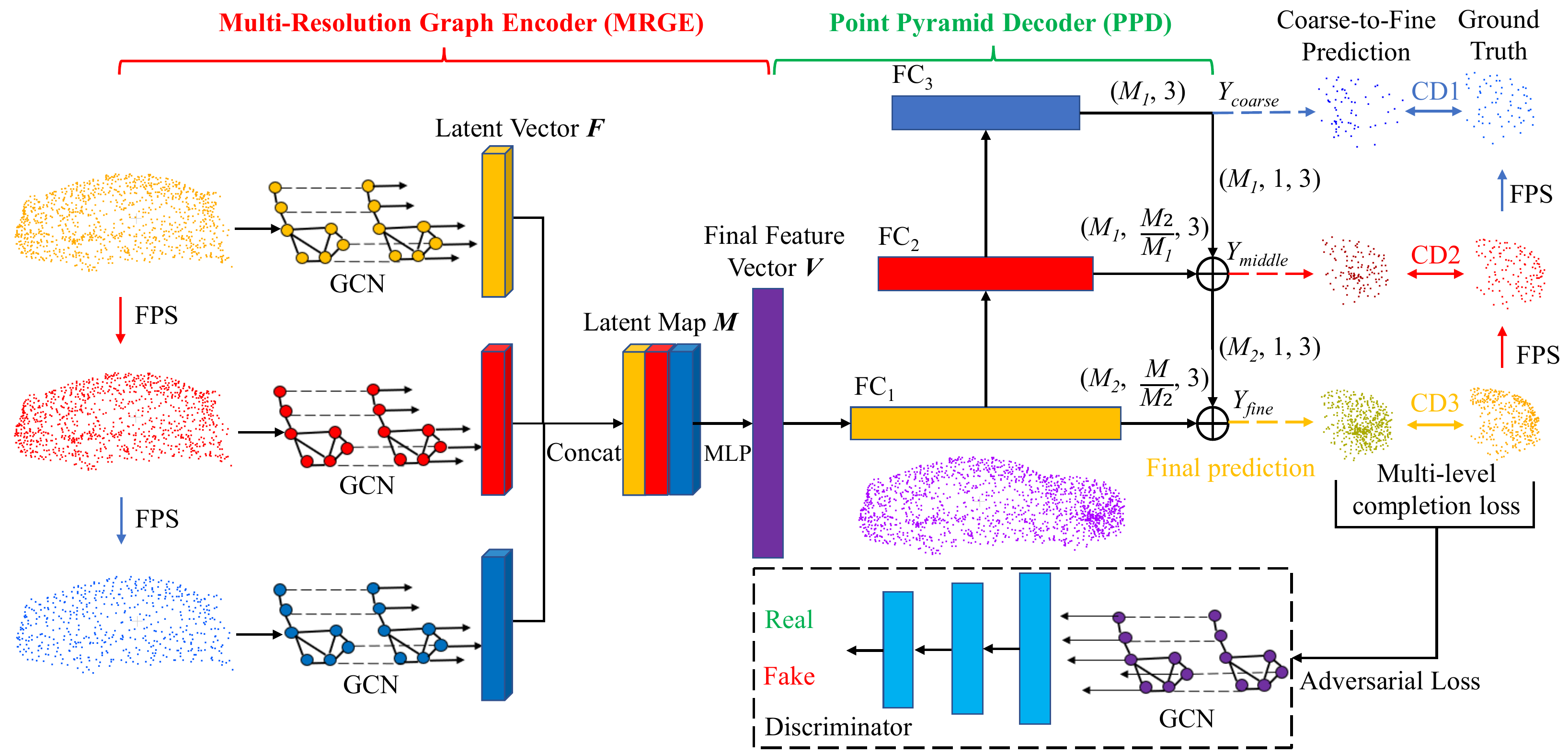} 
\caption{Architecture of the PC module. With input point clouds, it predicts additional parts of sparse and partial data by a Multi-Resolution Graph Encoder (MRGE) and a Point Pyramid Decoder (PPD). The Discriminator tries to distinguish the predicted regions from the real ones.}
\label{fig3}
\end{figure*}
\section{PC-RGNN}
In this section, we describe the proposed PC-RGNN in detail, including the entire framework as well as the modules of 3D proposal generation, point cloud completion, and attention based multi-scale graph feature aggregation.
\subsection{Framework}
The framework overview of our proposed PC-RGNN is illustrated in Fig. \ref{fig2}. The whole network consists of three main modules: (A) 3D proposal generation, (B) point cloud completion, and (C) attention based multi-scale graph neural network representation. Given a raw point cloud, the proposal generation module segments the foreground from background and generates 3D bounding box candidates simultaneously. The point cloud completion module then recovers dense and entire 3D shapes from sparse and/or partial proposal point clouds. Finally, the GNN module comprehensively encodes the structure characteristics to predict detection results.
\subsection{3D Proposal Generation}
As described in \cite{shi2019pointrcnn}, objects in 3D scenes are naturally separated, and the segmentation masks can be directly acquired from their 3D bounding box annotations. Therefore, we follow PointRCNN and build a sub-network to learn point-wise features to simultaneously locate the foreground areas and generate 3D proposals. Based on this bottom-up mechanism, we avoid using a large number of predefined 3D anchors and thus dramatically reduce the searching space in this phase.

Concretely, PointNet++ \cite{qi2017pointnet++} with multi-scale grouping is adopted as the backbone, and a segmentation head and a regression head are added to estimate the foreground mask and generate bounding box candidates respectively. For large outdoor scenes, the number of background points is usually much larger than that of foreground, and we therefore use the focal loss \cite{lin2017focal} in segmentation to deal with the class imbalance problem as:
\begin{equation}
\begin{array}{l}
{L_{{\rm{seg}}}}({p_t}) =  - {\alpha _t}{(1 - {p_t})^\gamma }\log ({p_t}),\vspace{1ex}\\
{\rm{where }}\;{p_t} = \left\{ \begin{array}{l}
p\qquad     {\rm{ for\;foreground\;points,}}\\
1 - p\quad{\rm{    otherwise}}{\rm{.}}
\end{array} \right.
\end{array}
\end{equation}
During training, we set ${\alpha _t} = 0.25$ and $\gamma  = 2$. For proposal generation, box locations are only regressed from foreground points. Here, a 3D bounding box is described as (\emph{x}, \emph{y}, \emph{z}, \emph{h}, \emph{w}, \emph{l}, $\theta $) in the LiDAR coordinate, where (\emph{x}, \emph{y}, \emph{z}) is the object center, (\emph{h}, \emph{w}, \emph{l}) is the object size, and $\theta $ is the object orientation from the bird’s eye view. For (\emph{z}, \emph{h}, \emph{w}, \emph{l}), the smooth L1 loss is utilized and for (\emph{x}, \emph{y}, $\theta $), the bin-based loss \cite{shi2020points} is exploited.
\subsection{Point Cloud Completion}
\begin{figure*}[t]
\centering
\includegraphics[width=0.97\textwidth]{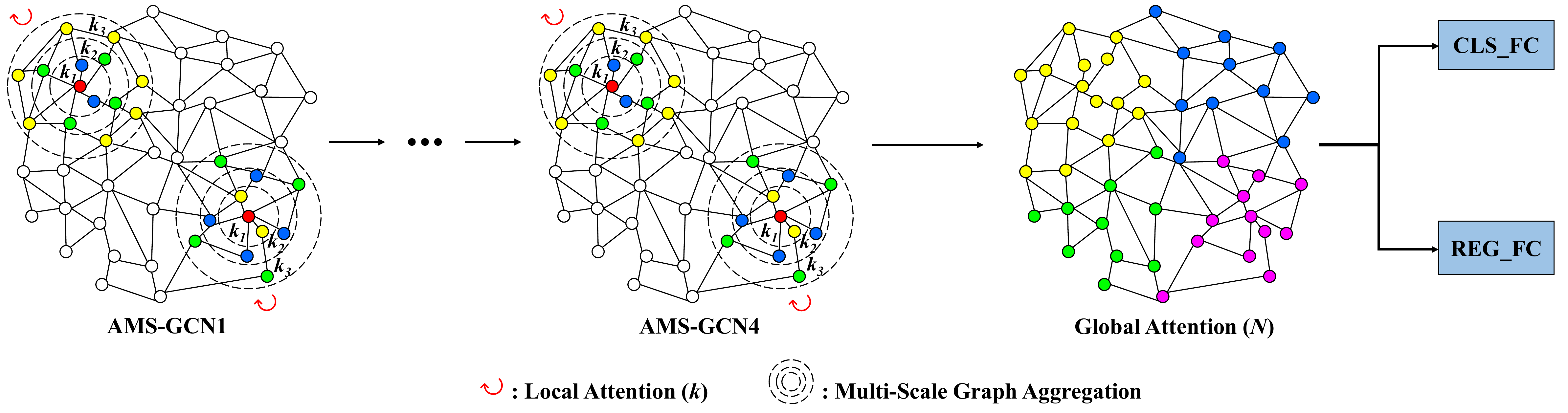} 
\caption{Architecture of the Attention based Multi-Scale Graph Neural Network.}
\label{fig4}
\end{figure*}
To address the challenges of sparse and partial data caused by distant and occluded objects, we propose a point cloud completion (PC) module for 3D detection. Unlike the existing 3D shape generation or reconstruction methods that output totally new point clouds, the proposed module only renders additional points as supplement with input data unchanged, aiming to preserve original spatial arrangement. As shown in Fig. \ref{fig3}, the overall architecture of the PC module is composed of three fundamental building blocks, \emph{i.e.} Multi-Resolution Graph Encoder (MRGE), Point Pyramid Decoder (PPD), and Discriminator Network.

The input to MRGE is an \emph{N} × 3 unordered point cloud. It is first down sampled to obtain two more views of smaller resolutions by farthest point sampling (FPS). Three independent GCN layers \cite{wang2019dynamic} are then used to map those resolutions into individual latent vectors \emph{F$_i$}. Compared with the PointNet-like models, GCN captures extra geometry clues from connection relations of points, which is expected to facilitate low-quality point cloud refinement. All the \emph{F$_i$} are further concatenated to form a stronger feature map \emph{M} in the size of 1920×3 and the feature maps are integrated into a final vector \emph{V}. Inspired by Feature Pyramid Networks \cite{lin2017feature}, PPD conducts in a coarse to fine fashion. Three feature layers FC$_1$, FC$_2$, FC$_3$ (size: 1024, 512, 256) are computed by passing \emph{V} through fully-connected layers. Each feature layer is responsible for recovering point clouds in a specific resolution. The coarse center points \emph{Y$_{coarse}$} are predicted from FC$_3$, which are of the size of \emph{M$_1$} × 3. The relative coordinates of middle center points \emph{Y$_{middle}$} are predicted from FC$_2$. Each point in \emph{Y$_{coarse}$} serves as the center to generate \emph{M$_2$}/\emph{M$_1$} points of \emph{Y$_{coarse}$}. Thus, the size of \emph{Y$_{middle}$} is \emph{M$_2$} × 3. Fine points \emph{Y$_{fine}$} are finally predicted by PPD, with the size of \emph{M} × 3.

The loss function is made up of two parts: the multi-level completion loss and the adversarial loss. The Chamfer Distance (CD) is chosen as the completion loss:
\begin{equation}
\begin{array}{l}
{d_{CD}}({S_1},{S_2}) = \frac{1}{{{S_1}}}\sum\limits_{x \in {S_1}} {\mathop {\min }\limits_{y \in {S_2}} } \left\| {x - y} \right\|_2^2\vspace{1ex}\\
{\kern 1pt} {\kern 1pt} {\kern 1pt} {\kern 1pt} {\kern 1pt} {\kern 1pt} {\kern 1pt} {\kern 1pt} {\kern 1pt} {\kern 1pt} {\kern 1pt} {\kern 1pt} {\kern 1pt} {\kern 1pt} {\kern 1pt} {\kern 1pt} {\kern 1pt} {\kern 1pt} {\kern 1pt} {\kern 1pt} {\kern 1pt} {\kern 1pt} {\kern 1pt} {\kern 1pt} {\kern 1pt} {\kern 1pt} {\kern 1pt} {\kern 1pt} {\kern 1pt} {\kern 1pt} {\kern 1pt} {\kern 1pt} {\kern 1pt} {\kern 1pt} {\kern 1pt} {\kern 1pt} {\kern 1pt} {\kern 1pt} {\kern 1pt} {\kern 1pt} {\kern 1pt} {\kern 1pt} {\kern 1pt} {\kern 1pt} {\kern 1pt} {\kern 1pt} {\kern 1pt} {\kern 1pt} {\kern 1pt} {\kern 1pt} {\kern 1pt} {\kern 1pt} {\kern 1pt} {\kern 1pt} {\kern 1pt} {\kern 1pt} {\kern 1pt} {\kern 1pt} {\kern 1pt} {\kern 1pt} {\kern 1pt} {\kern 1pt}  + \frac{1}{{{S_2}}}\sum\limits_{y \in {S_2}} {\mathop {\min }\limits_{x \in {S_1}} } \left\| {y - x} \right\|_2^2
\end{array}
\end{equation}
It measures the average nearest squared distance between the predicted point set \emph{S$_1$} and the ground truth \emph{S$_2$}. Since PPD estimates three point clouds at different levels, the multi-level completion loss is calculated in (3), where \emph{d}$_{CD1}$, \emph{d}$_{CD2}$, and \emph{d}$_{CD3}$ are weighted by a hyperparameter $\alpha$:
\begin{equation}
\begin{array}{l}
{L_{com}} = \alpha {d_{CD1}}({Y_{coarse}},Y_{gt}^{'})\vspace{1ex} + 2\alpha {d_{CD2}}({Y_{middle}},Y_{gt}^{''})\\
{\kern 1pt} {\kern 1pt} {\kern 1pt} {\kern 1pt} {\kern 1pt} {\kern 1pt} {\kern 1pt} {\kern 1pt} {\kern 1pt} {\kern 1pt} {\kern 1pt} {\kern 1pt} {\kern 1pt} {\kern 1pt} {\kern 1pt} {\kern 1pt} {\kern 1pt} {\kern 1pt} {\kern 1pt} {\kern 1pt} {\kern 1pt} {\kern 1pt} {\kern 1pt} {\kern 1pt} {\kern 1pt} {\kern 1pt} {\kern 1pt} {\kern 1pt} {\kern 1pt} {\kern 1pt} {\kern 1pt}  + {d_{CD3}}({Y_{fine}},{Y_{gt}})
\end{array}
\end{equation}
\begin{table*}[t]
\centering
\resizebox{2.0\columnwidth}{!}{
\begin{tabular}{l|c|ccc|ccc|ccc}
\hline
\multirow{2}{*}{Method} &\multirow{2}{*}{Modality} &\multicolumn{3}{|c|}{3D Object Detection (\%)} &\multicolumn{3}{|c|}{Bird's Eye View Detection (\%)} & \multicolumn{3}{|c}{Orientation Estimation (\%)}\\
\cline{3-11}
 &&Easy &Moderate &Hard &Easy &Moderate &Hard &Easy &Moderate &Hard\\
\hline
MV3D~\cite{chen2017multi} &RGB+LiDAR	&74.97	&63.63	&54.00	&86.62	&78.93	&69.80	&--	&--	&--\\
ContFuse~\cite{liang2018deep} &RGB+LiDAR	&83.68	&68.78	&61.67	&94.07	&85.35	&75.88	&--	&--	&--\\
AVOD-FPN~\cite{ku2018joint} &RGB+LiDAR	&83.07	&71.76	&65.73	&90.99	&84.82	&79.62	&94.65	&88.61	&83.71\\
F-PointNet~\cite{qi2018frustum} &RGB+LiDAR &82.19 &69.79	&60.59	&91.17	&84.67	&74.77	&--	&--	&--\\
MMF~\cite{liang2019multi} &RGB+LiDAR &88.40 &77.43 &70.22	&93.67	&88.21	&81.99	&--	&--	&--\\
\hline
VoxelNet~\cite{zhou2018voxelnet} &LiDAR only &77.47	&65.11	&57.73	&89.35	&79.26	&77.39	&--	&--	&--\\
SECOND~\cite{yan2018second} &LiDAR only	&83.13	&73.66	&66.20	&88.07	&79.37	&77.95	&87.84	&81.31	&71.95\\
PointPillars~\cite{lang2019pointpillars} &LiDAR only &82.58	&74.31	&68.99	&90.07	&86.56	&82.81	&93.84	&90.70	&87.47\\
Fast Point R-CNN~\cite{chen2019fast} &LiDAR only &85.29	&77.40	&70.24	&90.87	&87.84	&80.52	&--	&--	&--\\
STD~\cite{yang2019std}	&LiDAR only	&87.95	&79.71	&75.09	&94.74	&89.19	&86.42	&-- &--	&--\\
Part-A$^2$~\cite{shi2020points}	&LiDAR only	&87.81	&78.49	&73.51	&91.70	&87.79	&84.61	&95.00	&91.73	&88.86\\
3DSSD~\cite{yang20203dssd} &LiDAR only &88.36	&79.57	&74.55	&92.66 	&89.02	&85.86	&--	&--	&--\\
\hline
Point-GNN~\cite{shi2020point} &LiDAR only &88.33	&79.47	&72.29	&93.11 	&89.17	&83.90	&--	&--	&--\\
\textbf{\emph{Improvement}}	&LiDAR only	&\textbf{\emph{+0.80}}	&\textbf{\emph{+0.43}}	&\textbf{\emph{+3.25}}	&\textbf{\emph{+1.80}}	&\textbf{\emph{+0.45}}	&\textbf{\emph{+2.67}}	&--	&--	&--\\
\hline
PointRCNN~\cite{shi2019pointrcnn} &LiDAR only &86.96 &75.64	&70.70	&92.13	&87.39	&82.72	&95.90	&91.77	&86.92\\
\textbf{\emph{Improvement}}	&LiDAR only	&\textbf{\emph{+2.17}}	&\textbf{\emph{+4.26}}	&\textbf{\emph{+4.84}}	&\textbf{\emph{+2.78}}	&\textbf{\emph{+2.23}}	&\textbf{\emph{+3.85}}	&\textbf{\emph{+0.67}}	&\textbf{\emph{+1.46}}	&\textbf{\emph{+2.12}}\\
\hline
PC-RGNN (\textbf{Ours}) &LiDAR only  &\textbf{89.13} &\textbf{79.90} &\textbf{75.54}	&\textbf{94.91}	&\textbf{89.62}	&\textbf{86.57}	&\textbf{96.57}	&\textbf{93.23}	&\textbf{89.04}\\
\hline
\end{tabular}
}
\caption{Performance comparison in 3D object detection with previous state-of-the-art methods in terms of the car class on the KITTI test split (the results are computed by the official test server). 3D object detection and bird’s eye view detection are evaluated by Average Precision (AP) with IoU threshold 0.7, while orientation estimation is validated by Average Orientation Similarity (AOS) as mentioned in \cite{geiger2012we}.}
\label{table1}
\end{table*}
We define $F:X \to {Y^{'}}$ , which maps the low-quality input \emph{X} into the predicted additional point set ${Y^{'}}$. Discriminator \emph{D} tries to distinguish ${Y^{'}}$ from the real point set \emph{Y}. We first obtain a latent vector \emph{F} after two GCN layers and \emph{F} is then passed through fully-connected layers [256, 128, 16, 1] followed by a sigmoid-classifier to calculate the predicted value. In this case, the adversarial loss is defined as:
\vspace{1ex}\begin{equation}
{L_{adv}} = \sum\limits_{1 \le i \le S} {\log (D({y_i})} ) + \sum\limits_{1 \le i \le S} {\log (1 - D(F({x_i}))} )
\end{equation}
where ${x_i} \in X$, ${y_i} \in Y$, and \emph{S} is the dataset size. The total loss is thus formulated as:
\begin{equation}
{L_{com}} = {\lambda _{com}}{L_{com}} + {\lambda _{adv}}{L_{adv}}
\end{equation}
where ${\lambda _{com}}$ and ${\lambda _{adv}}$ are the weights of the completion loss and the adversarial loss.
\subsection{Attention based Multi-Scale GNN}
To comprehensively encode shape characteristics of point clouds in proposals refined by the PC module, we design a novel graph neural network module. It strengths the features delivered by the previous GNN counterpart by multi-scale contextual clue extraction and attention based discriminative point highlighting, thus named AMS-GNN. As demonstrated in Fig. \ref{fig4}, it is composed of four attention based multi-scale graph convolution (AMS-GCN) layers and a global attention (GA) layer. Each AMS-GCN layer contains a multi-scale graph aggregation operation and a local attention (LA) operation.

Specifically, we encode each proposal point cloud in a graph by regarding the points as vertices and the connection between points as edges, which makes features flow between neighbors. We define a point cloud as a set \emph{V} = \{\emph{v$_1$}, ..., \emph{v$_i$}, ..., \emph{v$_N$}\}, where \emph{v$_i$} = (\emph{p$_i$}, \emph{s$_i$}) is a point with both the 3D coordinates ${p_i} \in {R^3}$ and the state value ${s_i} \in {R^c}$, a \emph{c}-length vector that represents the point property. Given \emph{V}, we construct a graph \emph{G} = (\emph{V}, \emph{E}) by taking \emph{V} as the vertices and connecting each point to its \emph{k} neighbors as the edges \emph{E}.

Compared to the point cloud classification and segmentation tasks, point cloud detection is more complex as it regresses object positions and the points in different locations non-equally contribute to the results. Therefore, unlike the typical graph model preliminarily attempted by \cite{wang2019dynamic}, to dynamically adapt to the geometric structure of the object, we automatically learn the weight of each neighbor node when aggregating edge features. To this end, we design a GNN to extend the states of the vertices to include explicit position information and introduce an attention mechanism to assign individual weights to different nodes:
\begin{equation}
\begin{array}{l}
\Delta {p_i}^t = MLP_1^t(s_i^t)\vspace{1ex}\\
e_{ij}^t = MLP_2^t(concat({p_j} - {p_i} + \Delta {p_i}^t,s_j^t - s_i^t,s_i^t))\vspace{1ex}\\
\alpha _{ij}^t = softmax(MLP_3^t(e_{ij}^t)) = \frac{{exp(MLP_3^t(e_{ij}^t))}}{{\sum\nolimits_{k \in {N_i}} {exp(MLP_3^t(e_{ik}^t))} }}\vspace{1ex}\\
s_i^{t + 1} = sum(\alpha _{ij}^te_{ij}^t\left| {{\rm{ }}(i,j) \in E} \right.)
\end{array}
\end{equation}
where $s_i^t$, $s_j^t$ are the vertex features of the current node and its connecting node, respectively. MLP$_1$, MLP$_2$ and MLP$_3$ are used to learn position offset $\Delta {p_i}^t$, edge feature $e_{ij}^t$ and edge weight $\alpha _{ij}^t$ respectively.

In addition, for 3D object detection, multiple instances belonging to the same category often have different point cloud discretization distributions, which makes the features learned by graph nodes sensitive to graph resolution and connection relationship. To alleviate this interference, we aggregate multi-scale graph contextual features, and besides the local attention applied to the neighborhood, we also employ an attention to weight the global feature after four AMS-GCNs, which is called global attention. This local-global attention mechanism substantially makes the feature more powerful in detection.
\section{Experiments}
In this section, we subsequently present experimental evaluation, containing datasets and implementation details, results, and ablation studies.
\subsection{Datasets and Implementation Details}
We evaluate our PC-RGNN on the KITTI 3D object detection benchmark \cite{geiger2013vision} which contains 7481 training point clouds and 7518 testing ones. For fair comparison, we follow the previous studies \cite{chen2017multi,qi2018frustum} to subdivide the original training data into a training set and a validation set, resulting in 3712 samples for training and 3769 for validation. Since the point cloud in KITTI does not have a complete object shape, we first use ShapeNetCars with 1824 samples derived from \cite{yuan2018pcn} to train our point cloud completion module. In this way, we incorporate prior knowledge of the car class. Considering the variations in point cloud distribution, we extract 2000 object point clouds located in the ground-truth boxes with more than 2048 points from the KITTI training split to finetune the PC module. The ground-truth point cloud is created by uniformly sampling 2048 points on each shape. The low-quality point cloud is generated by randomly selecting a viewpoint and removing the points within a certain radius from original data. During training, each point cloud is transformed into the bounding box’s coordinates and projected back to the world frame after completion.

Our PC-RGNN is trained on 4 GTX 1080Ti GPUs using PyTorch. The stage-1 sub-network is trained for 200 epochs with the batch size at 16 and the learning rate of 0.002. The PC module is first pretrained for 60 epochs by using the Adam optimizer with an initial learning rate of 0.0001 and a batch size of 32. Combining the PC and AMS-GNN modules, the stage-2 sub-network is trained for 80 epochs with the batch size at 8 and the learning rate of 0.002 in an end-to-end manner.
\subsection{Results}
In the KITTI dataset, all the samples are divided into three sets with increasing difﬁculties, \emph{i.e.} Easy, Moderate and Hard, according to different bounding box heights and occlusion/truncation levels. For example, the occlusion levels in the three difﬁculties are `Fully visible', `Partly occluded', and `Difﬁcult to see', respectively.

Firstly, we evaluate PC-RGNN on the test set by submitting detection results to the official server. The results are summarized in Table \ref{table1}. PC-RGNN significantly outperforms the previously published state-of-the-art counterparts in all the tasks and difficulties. Point-GNN \cite{shi2020point} is the pioneer graph-based detector, and PointRCNN \cite{shi2019pointrcnn} is a representative two-stage approach. They are related to our method and compared to them, PC-RGNN makes two improvements at the second stage, \emph{i.e.} point cloud refinement by the PC module and feature enhancement by the AMS-GNN module. Therefore, we choose the two methods as the baselines. The performance gains delivered by PC-RGNN are emphasized in slanted bold font which indicate that our method is effective in LiDAR-based 3D object detection in the challenging scenes.
\begin{table}[t]
\centering
\resizebox{0.95\columnwidth}{!}{
\begin{tabular}{l|ccc}
\hline
\multirow{2}{*}{Method} &\multicolumn{3}{|c}{AP$_{3D}$ (IoU=0.7) (\%)}\\
\cline{2-4}
 &Easy &Moderate &Hard\\
\hline
MV3D~\cite{chen2017multi} 	&71.29	&62.68	&56.56\\
ContFuse~\cite{liang2018deep} 	&86.32	&73.25	&67.81\\
AVOD-FPN~\cite{ku2018joint} 	&84.41	&74.44	&68.65\\
F-PointNet~\cite{qi2018frustum}  &83.76	&70.92	&63.65\\
\hline
VoxelNet~\cite{zhou2018voxelnet}  &81.98	&65.46	&62.85\\
SECOND~\cite{yan2018second} 	&87.43	&76.48	&69.10\\
Fast Point R-CNN~\cite{chen2019fast}  &89.12	&79.00	&77.48\\
STD~\cite{yang2019std}		&89.70	&79.80	&79.30\\
Part-A$^2$~\cite{shi2020points}		&89.47	&79.47	&78.54\\
3DSSD~\cite{yang20203dssd}	&89.71	&79.45	&78.67\\
\hline
PointRCNN~\cite{shi2019pointrcnn}  &88.88	&78.63	&77.38\\
PC-RGNN (\textbf{Ours})   &\textbf{90.94} &\textbf{81.43} &\textbf{80.45}	\\
\textbf{\emph{Improvement}}		&\textbf{\emph{+2.06}}	&\textbf{\emph{+2.80}}	&\textbf{\emph{+3.07}}	\\
\hline
\end{tabular}
}
\caption{Performance comparison with previous state-of-the-art methods in terms of the car class on the KITTI val split.}
\label{table2}
\end{table}

\begin{figure*}[t]
\centering
\includegraphics[width=0.97\textwidth]{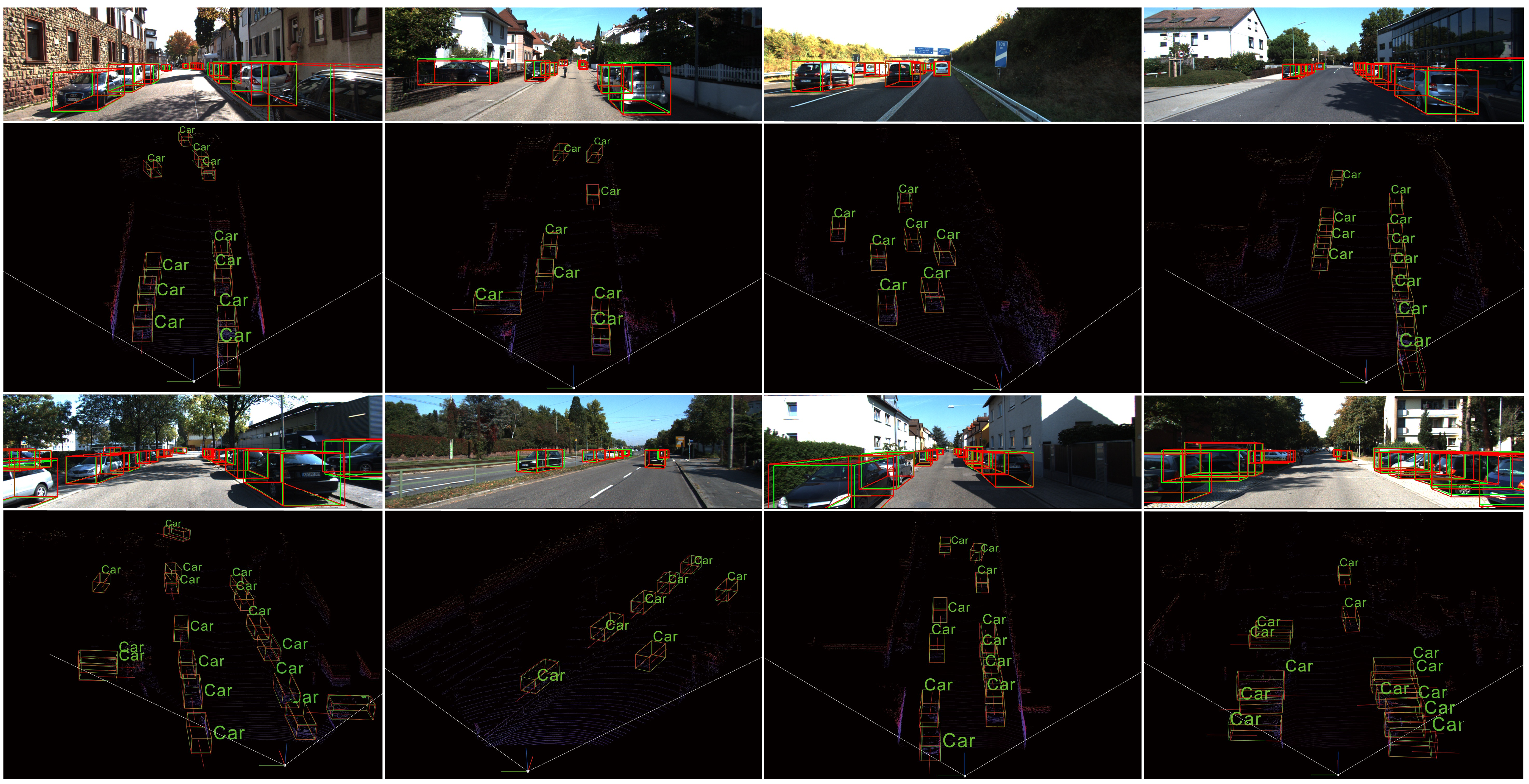} 
\caption{Qualitative results of PC-RGNN on the KITTI val split (best viewed with zoom-in). For each sample, the upper part is the image and the lower part is a representative view of the corresponding point cloud. The red boxes indicate the predicted results while the green ones denote the ground-truths.}
\label{fig5}
\end{figure*}

We then report the performance achieved on the KITTI validation set in Table \ref{table2}. We follow the official KITTI evaluation protocol, where the IoU threshold is 0.7 for the car class. The proposed PC-RGNN also outperforms all the other approaches by remarkable margins, especially in the Moderate and Hard difﬁculties. We present several challenging 3D detection examples in Fig. \ref{fig5}. We can observe that even in very difficult cases with distant and occluded objects, our network still reaches decent results, thanks to point cloud completion based data refining and attention multi-scale GNN based feature strengthening.
\subsection{Ablation Study}
\begin{table}[t]
\centering
\resizebox{0.95\columnwidth}{!}{
\begin{tabular}{cc|ccc|c|c}
\hline
\multicolumn{2}{c|}{Module} &\multicolumn{5}{|c}{AP$_{3D}$ (IoU=0.7) (\%)}\\
\hline
PC	&AMS-GNN	&Easy	&Moderate	&Hard	&3D mAP	&Gain\\
\hline
 & &88.88	&78.63	&77.38	&81.63	&--\\
 &	\textbf{$\checkmark$}	&89.62	&79.56	&78.43	&82.54	&\textbf{$\uparrow$0.91}\\
\textbf{$\checkmark$} & &89.97	&80.28	&79.29	&83.18	&\textbf{$\uparrow$1.55}\\
\textbf{$\checkmark$} &\textbf{$\checkmark$} &90.94	&81.43	&80.45	&84.27	&\textbf{$\uparrow$2.64}\\
\hline
\end{tabular}
}
\caption{Ablation study on the proposed PC and AMS-GNN modules.}
\label{table3}
\end{table}
\begin{figure}[t]
\centering
\includegraphics[width=0.95\columnwidth]{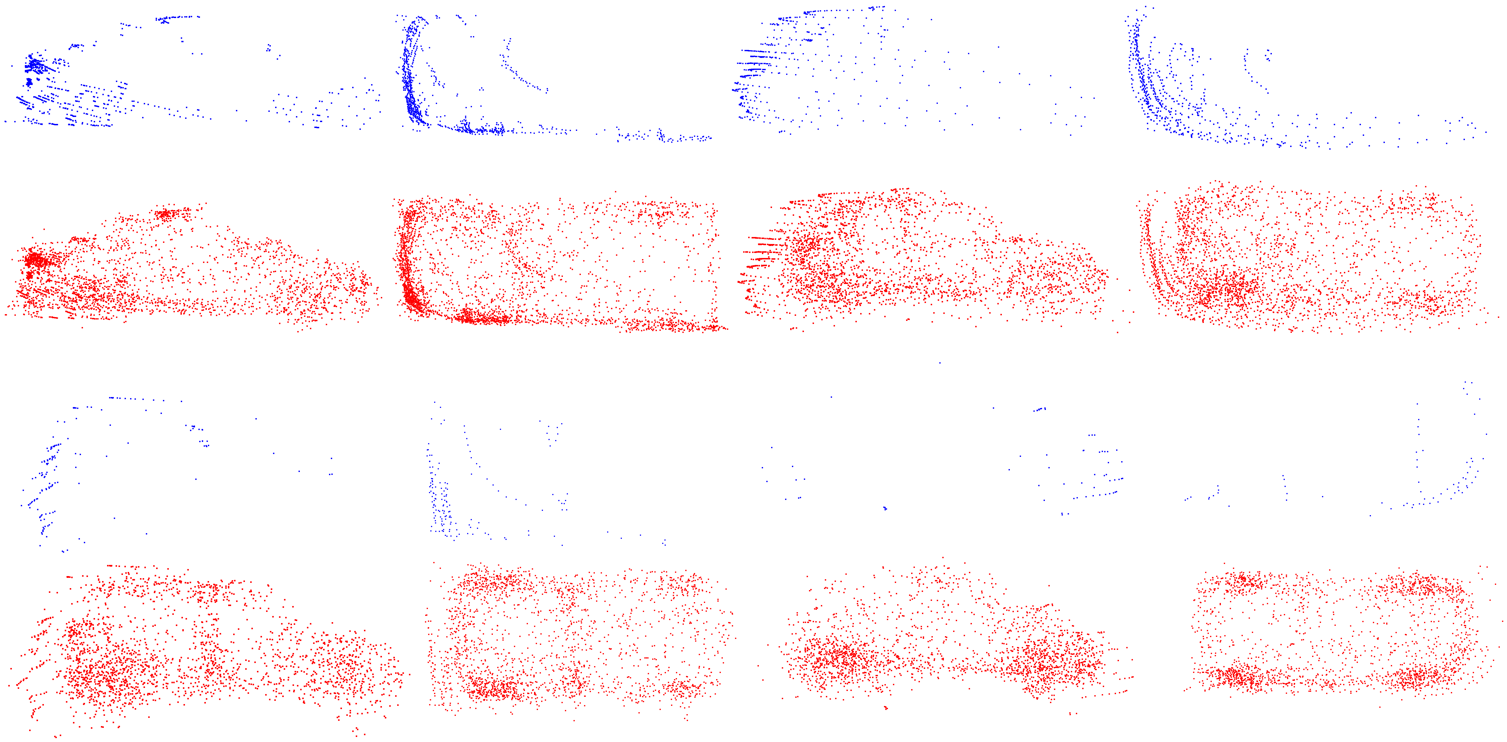} 
\caption{Point cloud completion results delivered by the PC module. Blue and red point clouds represent the data before and after this phase, respectively.}
\label{fig6}
\end{figure}
To better verify our contributions, we conduct ablation studies on the KITTI validation set. We primarily investigate the impacts of our proposed PC module as well as the AMS-GNN module to the ﬁnal results. We remove the PC module from PC-RGNN and replace our AMS-GNN with PointNet++ as our baseline, which achieves a 3D mAP of 81.63\%. We then add the PC and AMS-GNN modules separately on the baseline for comparison. Finally, we combine both the two modules to update the scores.

The results are shown in Table \ref{table3}. Only with either the PC module or the AMS-GNN module, the performance is boosted to 82.54\% and 83.18\%, respectively. Further, when combining both of them, it yields an mAP of 84.27\%, largely superior to that of the baseline model by 2.64\%. It clearly demonstrates the credits of the PC and AMS-GNN modules in PC-RGNN. Additionally, we visualize the object proposals before and after point cloud completion. As shown in Fig. \ref{fig6}, the original cars are barely recognizable due to the low-quality of the input data (\emph{i.e.} long-distance and heavy occlusion). In contrast, the refined point clouds display much more reasonable geometric shapes. It confirms the fact that the point cloud completion operation significantly increases the accuracy of 3D object detection in difficult scenes.

To further analyze the effectiveness of AMS-GNN, we take a graph neural network with four typical graph convolution layers derived from \cite{wang2019dynamic} as our baseline. We successively add the multi-scale (MS) graph aggregation, the local attention (LA), and the global attention (GA) to the baseline. All the results are listed in Table \ref{table4}. The improvements of the three parts are 0.21\%, 0.62\%, 0.16\%, respectively. This shows that both the local-global attention mechanism and the multi-scale graph feature aggregation improve the discriminative power of GNN in capturing geometric characteristics of point clouds.
\begin{table}[t]
\centering
\resizebox{0.95\columnwidth}{!}{
\begin{tabular}{ccc|ccc|c|c}
\hline
\multicolumn{3}{c|}{Module} &\multicolumn{5}{|c}{AP$_{3D}$ (IoU=0.7) (\%)}\\
\hline
MS	&LA	&GA	&Easy	&Moderate	&Hard	&3D mAP	&Gain\\
\hline
 & & &90.02	&80.39	&79.44	&83.28	&--\\
\textbf{$\checkmark$} & & &90.16 &80.57	&79.73 &83.49 &\textbf{$\uparrow$0.21}\\
\textbf{$\checkmark$} &\textbf{$\checkmark$} & &90.78 &81.25 &80.29	&84.11 &\textbf{$\uparrow$0.83}\\
\textbf{$\checkmark$} &\textbf{$\checkmark$} &\textbf{$\checkmark$} &90.94	&81.43	&80.45	&84.27	&\textbf{$\uparrow$0.99}\\
\hline
\end{tabular}
}
\caption{Ablation experiments on the AMS-GNN module.}
\label{table4}
\end{table}
\begin{table}[t]
\centering
\resizebox{0.95\columnwidth}{!}{
\begin{tabular}{c|ccc|c|c}
\hline
\multirow{2}{*}{Method} &\multicolumn{5}{|c}{AP$_{3D}$ (IoU=0.7) (\%)}\\
\cline{2-6}
 &Easy	&Moderate	&Hard	&3D mAP	&Gain\\
\hline
PC-M &90.94	&81.43	&80.45	&84.27	&--\\
PC-M without GE	&90.72	&80.48	&79.81	&83.67	&\textbf{$\downarrow$0.60}\\
PC-M without MR &90.55	&80.60	&79.66	&83.60	&\textbf{$\downarrow$0.67}\\
PC-O &90.42	&80.51	&79.29	&83.41	&\textbf{$\downarrow$0.86}\\
\hline
\end{tabular}
}
\caption{Ablation experiments on the PC module.}
\label{table5}
\end{table}

Besides, we validate the PC module. We take the complete PC-RGNN as our baseline, and its point cloud completion module (PC-M) only predicts additional points for refinement according to low-quality input. We then replace the graph encoder in PC-M with PointNet and remove the multi-resolution branch, generating two new models named PC-M without GE and PC-M without MR, respectively.  Meanwhile, we change PC-M to the module which generates totally new shapes from low-quality input as existing point cloud completion networks do and name it as PC-O. As shown in Table \ref{table5}, compared with our baseline, PC-M without GE and PC-M without MR decrease by 0.60\% and 0.67\%, respectively. This proves that the multi-resolution graph encoding strategy indeed captures additional geometric features. Compared with the baseline, PC-O encounters a drop of 0.86\%. It suggests that despite the low-quality, the original points are critical to regress the object locations, and the way proposed in this study, \emph{i.e.} PC-M, well handles this problem.
\section{Conclusion}
This paper proposes a novel two-stage approach, namely PC-RGNN, to LiDAR-based 3D object detection. It aims to address low-quality inputs, \emph{i.e.} sparsely and partially sampled point clouds, caused by distant and/or occluded objects in challenging scenarios. To this end, we introduce a point cloud completion module for data refinement, and to the best of our knowledge, this is the first attempt to integrate this technique into 3D object detection framework. Furthermore, we design a new graph neural network for feature enhancement, which comprehensively captures the geometric relations among points by a local-global attention mechanism and multi-scale graph based contextual information aggregation. Extensive experiments are carried out on the KITTI dataset and state of the art results are reached, which demonstrate the effectiveness of the proposed PC-RGNN.
\setcounter{secnumdepth}{0}
\section{Acknowledgment}
This work is partly supported by the National Natural Science Foundation of China (No. 62022011), the Research Program of State Key Laboratory of Software Development Environment (SKLSDE-2019ZX-03), and the Fundamental Research Funds for the Central Universities.
{\bibliography{references}\bibliographystyle{aaai}}

\begin{thebibliography}{28}
\providecommand{\natexlab}[1]{#1}
\providecommand{\url}[1]{\texttt{#1}}
\providecommand{\urlprefix}{URL }
\expandafter\ifx\csname urlstyle\endcsname\relax
  \providecommand{\doi}[1]{doi:\discretionary{}{}{}#1}\else
  \providecommand{\doi}{doi:\discretionary{}{}{}\begingroup
  \urlstyle{rm}\Url}\fi

\bibitem[{Achlioptas et~al.(2018)Achlioptas, Diamanti, Mitliagkas, and
  Guibas}]{achlioptas2018learning}
Achlioptas, P.; Diamanti, O.; Mitliagkas, I.; and Guibas, L. 2018.
\newblock Learning representations and generative models for 3d point clouds.
\newblock In \emph{International Conference on Machine Learning}, 40--49.

\bibitem[{Chen et~al.(2017)Chen, Ma, Wan, Li, and Xia}]{chen2017multi}
Chen, X.; Ma, H.; Wan, J.; Li, B.; and Xia, T. 2017.
\newblock Multi-view 3d object detection network for autonomous driving.
\newblock In \emph{IEEE Conference on Computer Vision and Pattern Recognition},
  1907--1915.

\bibitem[{Chen et~al.(2019)Chen, Liu, Shen, and Jia}]{chen2019fast}
Chen, Y.; Liu, S.; Shen, X.; and Jia, J. 2019.
\newblock Fast point r-cnn.
\newblock In \emph{IEEE International Conference on Computer Vision},
  9775--9784.

\bibitem[{Engelcke et~al.(2017)Engelcke, Rao, Wang, Tong, and
  Posner}]{engelcke2017vote3deep}
Engelcke, M.; Rao, D.; Wang, D.~Z.; Tong, C.~H.; and Posner, I. 2017.
\newblock Vote3deep: Fast object detection in 3d point clouds using efficient
  convolutional neural networks.
\newblock In \emph{IEEE International Conference on Robotics and Automation},
  1355--1361.

\bibitem[{Geiger et~al.(2013)Geiger, Lenz, Stiller, and
  Urtasun}]{geiger2013vision}
Geiger, A.; Lenz, P.; Stiller, C.; and Urtasun, R. 2013.
\newblock Vision meets robotics: The kitti dataset.
\newblock \emph{The International Journal of Robotics Research} 32(11):
  1231--1237.

\bibitem[{Geiger, Lenz, and Urtasun(2012)}]{geiger2012we}
Geiger, A.; Lenz, P.; and Urtasun, R. 2012.
\newblock Are we ready for autonomous driving? the kitti vision benchmark
  suite.
\newblock In \emph{IEEE Conference on Computer Vision and Pattern Recognition},
  3354--3361.

\bibitem[{Huang et~al.(2020)Huang, Yu, Xu, Ni, and Le}]{huang2020pf}
Huang, Z.; Yu, Y.; Xu, J.; Ni, F.; and Le, X. 2020.
\newblock PF-Net: Point fractal network for 3D point cloud completion.
\newblock In \emph{IEEE Conference on Computer Vision and Pattern Recognition},
  7662--7670.

\bibitem[{Ku et~al.(2018)Ku, Mozifian, Lee, Harakeh, and
  Waslander}]{ku2018joint}
Ku, J.; Mozifian, M.; Lee, J.; Harakeh, A.; and Waslander, S.~L. 2018.
\newblock Joint 3d proposal generation and object detection from view
  aggregation.
\newblock In \emph{IEEE International Conference on Intelligent Robots and
  Systems}, 1--8.

\bibitem[{Lang et~al.(2019)Lang, Vora, Caesar, Zhou, Yang, and
  Beijbom}]{lang2019pointpillars}
Lang, A.~H.; Vora, S.; Caesar, H.; Zhou, L.; Yang, J.; and Beijbom, O. 2019.
\newblock Pointpillars: Fast encoders for object detection from point clouds.
\newblock In \emph{IEEE Conference on Computer Vision and Pattern Recognition},
  12697--12705.

\bibitem[{Liang et~al.(2019)Liang, Yang, Chen, Hu, and
  Urtasun}]{liang2019multi}
Liang, M.; Yang, B.; Chen, Y.; Hu, R.; and Urtasun, R. 2019.
\newblock Multi-task multi-sensor fusion for 3d object detection.
\newblock In \emph{IEEE Conference on Computer Vision and Pattern Recognition},
  7345--7353.

\bibitem[{Liang et~al.(2018)Liang, Yang, Wang, and Urtasun}]{liang2018deep}
Liang, M.; Yang, B.; Wang, S.; and Urtasun, R. 2018.
\newblock Deep continuous fusion for multi-sensor 3d object detection.
\newblock In \emph{European Conference on Computer Vision}, 641--656.

\bibitem[{Lin et~al.(2017{\natexlab{a}})Lin, Doll{\'a}r, Girshick, He,
  Hariharan, and Belongie}]{lin2017feature}
Lin, T.-Y.; Doll{\'a}r, P.; Girshick, R.; He, K.; Hariharan, B.; and Belongie,
  S. 2017{\natexlab{a}}.
\newblock Feature pyramid networks for object detection.
\newblock In \emph{IEEE Conference on Computer Vision and Pattern Recognition},
  2117--2125.

\bibitem[{Lin et~al.(2017{\natexlab{b}})Lin, Goyal, Girshick, He, and
  Doll{\'a}r}]{lin2017focal}
Lin, T.-Y.; Goyal, P.; Girshick, R.; He, K.; and Doll{\'a}r, P.
  2017{\natexlab{b}}.
\newblock Focal loss for dense object detection.
\newblock In \emph{IEEE Conference on Computer Vision and Pattern Recognition},
  2980--2988.

\bibitem[{Qi et~al.(2019)Qi, Litany, He, and Guibas}]{qi2019deep}
Qi, C.~R.; Litany, O.; He, K.; and Guibas, L.~J. 2019.
\newblock Deep hough voting for 3d object detection in point clouds.
\newblock In \emph{IEEE International Conference on Computer Vision},
  9277--9286.

\bibitem[{Qi et~al.(2018)Qi, Liu, Wu, Su, and Guibas}]{qi2018frustum}
Qi, C.~R.; Liu, W.; Wu, C.; Su, H.; and Guibas, L.~J. 2018.
\newblock Frustum pointnets for 3d object detection from rgb-d data.
\newblock In \emph{IEEE Conference on Computer Vision and Pattern Recognition},
  918--927.

\bibitem[{Qi et~al.(2017{\natexlab{a}})Qi, Su, Mo, and Guibas}]{qi2017pointnet}
Qi, C.~R.; Su, H.; Mo, K.; and Guibas, L.~J. 2017{\natexlab{a}}.
\newblock Pointnet: Deep learning on point sets for 3d classification and
  segmentation.
\newblock In \emph{IEEE Conference on Computer Vision and Pattern Recognition},
  652--660.

\bibitem[{Qi et~al.(2017{\natexlab{b}})Qi, Yi, Su, and
  Guibas}]{qi2017pointnet++}
Qi, C.~R.; Yi, L.; Su, H.; and Guibas, L.~J. 2017{\natexlab{b}}.
\newblock Pointnet++: Deep hierarchical feature learning on point sets in a
  metric space.
\newblock In \emph{Advances in Neural Information Processing Systems},
  5099--5108.

\bibitem[{Sarmad, Lee, and Kim(2019)}]{sarmad2019rl}
Sarmad, M.; Lee, H.~J.; and Kim, Y.~M. 2019.
\newblock Rl-gan-net: A reinforcement learning agent controlled gan network for
  real-time point cloud shape completion.
\newblock In \emph{IEEE Conference on Computer Vision and Pattern Recognition},
  5898--5907.

\bibitem[{Shi, Wang, and Li(2019)}]{shi2019pointrcnn}
Shi, S.; Wang, X.; and Li, H. 2019.
\newblock Pointrcnn: 3d object proposal generation and detection from point
  cloud.
\newblock In \emph{IEEE Conference on Computer Vision and Pattern Recognition},
  770--779.

\bibitem[{Shi et~al.(2020)Shi, Wang, Shi, Wang, and Li}]{shi2020points}
Shi, S.; Wang, Z.; Shi, J.; Wang, X.; and Li, H. 2020.
\newblock From points to parts: 3d object detection from point cloud with
  part-aware and part-aggregation network.
\newblock \emph{IEEE Transactions on Pattern Analysis and Machine Intelligence}
  .

\bibitem[{Shi and Rajkumar(2020)}]{shi2020point}
Shi, W.; and Rajkumar, R. 2020.
\newblock Point-gnn: Graph neural network for 3d object detection in a point
  cloud.
\newblock In \emph{IEEE Conference on Computer Vision and Pattern Recognition},
  1711--1719.

\bibitem[{Wang and Posner(2015)}]{wang2015voting}
Wang, D.~Z.; and Posner, I. 2015.
\newblock Voting for voting in online point cloud object detection.
\newblock In \emph{Robotics: Science and Systems}, volume~1, 10--15607.

\bibitem[{Wang et~al.(2019)Wang, Sun, Liu, Sarma, Bronstein, and
  Solomon}]{wang2019dynamic}
Wang, Y.; Sun, Y.; Liu, Z.; Sarma, S.~E.; Bronstein, M.~M.; and Solomon, J.~M.
  2019.
\newblock Dynamic graph cnn for learning on point clouds.
\newblock \emph{ACM Transactions on Graphics} 38(5): 1--12.

\bibitem[{Yan, Mao, and Li(2018)}]{yan2018second}
Yan, Y.; Mao, Y.; and Li, B. 2018.
\newblock Second: Sparsely embedded convolutional detection.
\newblock \emph{Sensors} 18(10): 3337.

\bibitem[{Yang et~al.(2020)Yang, Sun, Liu, and Jia}]{yang20203dssd}
Yang, Z.; Sun, Y.; Liu, S.; and Jia, J. 2020.
\newblock 3dssd: Point-based 3d single stage object detector.
\newblock In \emph{IEEE Conference on Computer Vision and Pattern Recognition},
  11040--11048.

\bibitem[{Yang et~al.(2019)Yang, Sun, Liu, Shen, and Jia}]{yang2019std}
Yang, Z.; Sun, Y.; Liu, S.; Shen, X.; and Jia, J. 2019.
\newblock Std: Sparse-to-dense 3d object detector for point cloud.
\newblock In \emph{IEEE International Conference on Computer Vision},
  1951--1960.

\bibitem[{Yuan et~al.(2018)Yuan, Khot, Held, Mertz, and Hebert}]{yuan2018pcn}
Yuan, W.; Khot, T.; Held, D.; Mertz, C.; and Hebert, M. 2018.
\newblock Pcn: Point completion network.
\newblock In \emph{IEEE Conference on 3D Vision}, 728--737.

\bibitem[{Zhou and Tuzel(2018)}]{zhou2018voxelnet}
Zhou, Y.; and Tuzel, O. 2018.
\newblock Voxelnet: End-to-end learning for point cloud based 3d object
  detection.
\newblock In \emph{IEEE Conference on Computer Vision and Pattern Recognition},
  4490--4499.

\end{thebibliography}

\end{document}